\documentclass{article}

\usepackage{arxiv}

\usepackage[utf8]{inputenc} % allow utf-8 input
\usepackage[T1]{fontenc}    % use 8-bit T1 fonts
\usepackage{hyperref}       % hyperlinks
\usepackage{url}            % simple URL typesetting
\usepackage{booktabs}       % professional-quality tables
\usepackage{amsfonts}       % blackboard math symbols
\usepackage{nicefrac}       % compact symbols for 1/2, etc.
\usepackage{microtype}      % microtypography
\usepackage{lipsum}
\usepackage{graphicx}
\usepackage{xcolor}
\graphicspath{ {./images/} }

\newcommand{\datasetname}{PoseTrackReID}

\title{\datasetname: Dataset Description}

\author{
 Andreas Doering \\
   \And
 Di Chen \\
  \And
 Shanshan  Zhang \\
 \And 
 Bernt Schiele \\
 \And 
 Juergen Gall \\
}

\begin{document}
\maketitle
\begin{abstract}
Current datasets for video-based person re-identification~(re-ID) do not include structural knowledge in form of human pose annotations for the persons of interest. Nonetheless, pose information is very helpful to disentangle useful feature information from background or occlusion noise. Especially real-world scenarios, such as surveillance, contain a lot of occlusions in human crowds or by obstacles. On the other hand, video-based person re-ID can benefit other tasks such as multi-person pose tracking in terms of robust feature matching. 
For that reason, we present \datasetname, a large-scale dataset for multi-person pose tracking and video-based person re-ID.  With \datasetname, we want to bridge the gap between person re-ID and multi-person pose tracking. Additionally, this dataset provides a good benchmark for current state-of-the-art methods on multi-frame person re-ID.

\end{abstract}

% keywords can be removed
\keywords{Dataset \and Pose Estimation \and Multi-Person Pose Tracking \and Person re-identification}

\section{Introduction}
Multi-person pose tracking \cite{snower202015, wang2020combining, rafi2020selfsupervised} and video-based person re-identification (re-ID) \cite{hou2020temporal,yan2020learning,yang2020spatial,zhang2020multi} are two very active research fields in computer vision, 
relevant for many applications like sports, autonomous driving and security. Both tasks are challenging due to different viewpoints, varying (low) image resolutions, illumination changes, occlusions, motion blur and crowded scenes.

In multi-person pose tracking, the target is to estimate all human poses for all frames of a video and link them over time. Person re-identification, on the other hand, aims to find a person of interest across multiple sequences and non-overlapping cameras. Although both tasks are mainly studied separately, they can benefit from each other. For example, person re-identification can benefit from human pose estimation by utilizing the semantic structure of the person of interest \cite{Miao_2019_ICCV}. Person re-ID, on the other hand, can be utilized to link all instances of the same person across time \cite{zhang2019exploiting}.

Common datasets for person re-ID~\cite{7410490,zheng2016mars,zheng2017unlabeled,wu2018cvpr_oneshot} provide tight person crops and mostly contain the person of interest only. Though, in many real world scenarios such as surveillance or sports, persons are often occluded by obstacles. Further, these scenarios contain a lot of crowded scenes in which persons frequently occlude each other. This leads to a lot of ambiguities, especially if a person is partially visible or multiple persons are present within a single crop. 
% To overcome problems such as occlusion, Miao et al. \cite{Miao_2019_ICCV} propose the Occluded-DukeMTMC dataset as an extension of the DukeMTMC-ReID dataset \cite{wu2018cvpr_oneshot} and provide keypoint information obtained from an off-the-shelf pose estimator \cite{cao2017realtime}. 
To overcome problems such as occlusion, Miao et al.~\cite{Miao_2019_ICCV} propose the Occluded-DukeMTMC dataset as an extension of the DukeMTMC-ReID dataset \cite{zheng2017unlabeled} and provide keypoint information obtained from an off-the-shelf pose estimator \cite{cao2017realtime}. 
However, off-the-shelf pose estimators such as \cite{cao2017realtime} perform poorly in crowded scenes with severe occlusions, resulting in misplaced keypoints. 

Video keypoint annotations, on the other hand, are costly and time consuming. Consequently, a limited amount of training data is publicly available, especially for the task of multi-person pose tracking.
We argue that pose tracking and person re-ID are two related tasks which should be solved jointly. For that reason, we extended a large-scale multi-person pose estimation and tracking dataset (PoseTrack \cite{PoseTrack}) and propose \datasetname. \datasetname~combines both tasks of multi-person pose tracking and video-based person re-ID with the aim to improve pose-based person re-ID methods in videos. To the best of our knowledge, this is the first dataset combining person re-ID and multi-person pose tracking. 

\section{Dataset}
We briefly describe the PoseTrack 2018 dataset \cite{PoseTrack} in Section \ref{posetrack} and introduce \datasetname~as well as its annotation process in Section \ref{our_dataset}.

\subsection{PoseTrack}\label{posetrack}
The PoseTrack 2018 dataset\footnote{We refer to the final version of PoseTrack 2018 which was released in 2019, containing more sequences as described in \cite{PoseTrack}.} \cite{PoseTrack} is a large-scale dataset for multi-person pose estimation and tracking, containing 823 sequences. Following the split of the  MPII Human Pose dataset \cite{andriluka14cvpr}, the sequences are split into 293, 170 and 375 videos for training, validation and testing. The majority of sequences range from 41 to 151 frames and contain 30 densely annotated frames around the middle of each sequence. On top, validation and test sequences are annotated with a step size of four frames. In total, PoseTrack 2018 consists of 46,933 labeled frames.

PoseTrack provides ignore regions to exclude crowds and small people, keypoint annotations, head bounding boxes to estimate the scale of a person and track ids. Unfortunately, track ids are not unique throughout the dataset, not even within a single sequence. If a person leaves a scene and re-enters, for instance, it is assigned a new track id. This can result in ambiguities for appearance-based similarity approaches.

\subsection{\datasetname}\label{our_dataset}
The PoseTrack 2018 dataset contains multiple sequences sampled from the same video. For that reason, the dataset is well suited to be extended by person IDs as the same person appears in multiple sequences. In addition, PoseTrack contains a lot of challenging sequences with a high degree of occlusion and motion blur. In contrast to related person re-ID datasets \cite{7410490,zheng2016mars,zheng2017unlabeled,wu2018cvpr_oneshot}, this results in person bounding boxes with a high degree of occlusion and multiple persons within a single person crop. Due to its challenging nature, our proposed dataset can provide a new measure of robustness for current state-of-the-art approaches. The provided keypoint annotations further allow to incorporate structural pose information paving the way for pose-based person re-ID methods. On the other hand, our dataset opens ways for re-ID based pose tracking methods.
.
\subsection{Data Annotation}
We developed a new tool to annotate the PoseTrack 2018 dataset and adopt the same annotation format. %Unfortunately, the annotations for the test set are not publicly available. 
At the moment, we only provide annotations for the training and validation sets. 

The annotation process was performed in four steps. First, we re-annotated the existing bounding boxes on the training and validation set as the bounding boxes provided by PoseTrack have been an artifact of the annotation process and were therefore not accurately annotated. Specifically, most bounding boxes only contained the visible part of a person. We extended all bounding boxes to cover the entire portion of a person, including occluded body parts. 
To increase the difficulty of person re-ID, we further annotated bounding boxes for small persons and persons in crowds. Then, we interpolated and revised the bounding boxes between non-keyframes\footnote{We denote as keyframe all frames which were annotated in PoseTrack 2018.}. This results in additional \textbf{25,478} and \textbf{11,238} annotated frames on the training and validation sets, respectively.  Second, we adapted the ignore regions on the training set such that the newly added persons are not anymore excluded. 
%At the current state of our dataset, we removed the ignore regions on the validation set. In contrast to PoseTrack 2018, the validation set is not intended to be used for training.
In a third step, we annotated unique person identities throughout the training and validation sets. Additionally, we added head bounding boxes for newly added persons for all frames on the validation set. On the training set, we only provide head bounding boxes from PoseTrack 2018.

In a final step, we adapt and annotate person keypoints for all keyframes on the training and validation sets. The original annotations of the PoseTrack 2018 dataset consist of 15 keypoints where each keypoint contains a flag, whether it is annotated or not. Unfortunately, there exist several cases in which these flags are not set reliably. In our dataset, we re-define the purpose of the keypoint flags and include occluded keypoints. In that way, pose estimation and re-ID approaches can utilize occlusion information within their training pipelines. We define a joint $j = (x, y, v)$ as occluded, if $x \ge 0$, $y \ge 0$ and $v = 0$. A joint is truncated if $x = -1, y=-1$ and $v = 0$. Otherwise, a joint is defined as visible if $v = 1$.
After refining the original keypoints, we ran an off-the-shelf pose estimator \cite{rafi2020selfsupervised} to estimate the poses for all newly added bounding boxes on keyframes only, which were then refined manually afterwards. 

\bibliographystyle{unsrt}
\bibliography{references}

\end{document}